\documentclass{article} 

\usepackage{iclr2025_conference,times}  
\usepackage{amsmath,amssymb,amsfonts}
\usepackage{algorithmic}
\usepackage{graphicx}
\usepackage{textcomp}
\usepackage{xcolor}
\usepackage{url}
\usepackage{booktabs}
\usepackage{multirow}
\usepackage[utf8]{inputenc}
\usepackage{hyperref}
\usepackage[T1]{fontenc}
\usepackage{subcaption}

\title{Knowledge Graphs as World Models for Semantic Material-Aware Obstacle Handling in Autonomous Vehicles}

\author{Ayush Bheemaiah \& Seungyong Yang \\
Artificial Intelligence Explainability Accountability (AIEA) Lab \\
University of California, Santa Cruz \\
Santa Cruz, CA 95064, USA \\
\texttt{ayushbheemaiah@gmail.com, syyang26@pupils.nlcsjeju.kr} 
}

\iclrfinalcopy 
\begin{document}

\maketitle

\begin{abstract}
The inability of autonomous vehicles (AVs) to infer the material properties of obstacles limits their decision-making capacity. While AVs rely on sensor systems such as cameras, LiDAR, and radar to detect obstacles, this study suggests combining sensors with a knowledge graph (KG)-based world model to improve AVs' comprehension of physical material qualities. Beyond sensor data, AVs can infer qualities such as malleability, density, and elasticity using a semantic KG that depicts the relationships between obstacles and their attributes. Using the CARLA autonomous driving simulator, we evaluated AV performance with and without KG integration. The findings demonstrate that the KG-based method improves obstacle management, which allows AVs to use material qualities to make better decisions about when to change lanes or apply emergency braking. For example, the KG-integrated AV changed lanes for hard impediments like traffic cones and successfully avoided collisions with flexible items such as plastic bags by passing over them. Compared to the control system, the KG framework demonstrated improved responsiveness to obstacles by resolving conflicting sensor data, causing emergency stops for 13.3\% more cases. In addition, our method exhibits a 6.6\% higher success rate in lane-changing maneuvers in experimental scenarios, particularly for larger, high-impact obstacles. While we focus particularly on autonomous driving, our work demonstrates the potential of KG-based world models to improve decision-making in embodied AI systems and scale to other domains, including robotics, healthcare, and environmental simulation.
\end{abstract}

\section{Introduction}
Autonomous vehicles (AV) have gained rapid market penetration in recent years, yet the industry’s goal of achieving Level 5 automation remains extremely difficult, with forecasts suggesting that fully self-driving cars will not be available until 2035 \citep{Level_5}. The transition from current AVs at Level 3, which still require human supervision, to complete autonomy under any Operational Design Domain (ODD) remains distant \citep{Level_3}. Therefore, fundamental approaches to autonomous vehicle development must be reevaluated.

The current AVs available to the public highly depend on visual perception that mirrors human visual cognition through heterogeneous sensors such as cameras, LiDAR, and RADAR \citep{sensors}. However, this approach oversimplifies the complex decision-making processes involved in human driving, as human drivers integrate multiple layers of knowledge that transcend beyond visual and spatial recognition using sensors: (i) recognition of physical properties of obstacles, (ii) moral and emotional considerations, (iii) anticipatory behavior in driving (iv) contextual and situational understanding \citep{human_drivers}. 

AVs primarily show low performance in performing contextual decisions about common scenarios related to obstacle management, such as distinguishing the appropriate response to a lightweight plastic bag placed on the road \citep{obstacle}. Unlike human drivers, AVs fail to replicate the intuitive understanding required to navigate obstacles and fall short in understanding physical properties to optimize their decision-making and minimize the damage after a potential collision with an object. More specifically, lightweight objects like empty plastic bags can be safely driven over, while rigid paper boxes require lane changes. Hence, while human drivers can assess such obstacles by instinctively understanding physical properties, AVs cannot incorporate material characteristics into their reasoning \citep{material}. Thus, a comprehensive framework that amalgamates the recognition of material properties with existing heterogeneous sensor systems is fundamental for enabling AVs to handle obstacles that exceed the interpretative capabilities of existing sensors.

Human drivers, when faced with atypical situations such as a large plastic bag blowing across the road, rely on a combination of visual perception, situational judgment, and adaptive behavior to make their decisions \citep{atypical}. By accurately assessing the object's size, shape, texture, and movement patterns visually and intuitively, they evaluate its potential threat, allowing them to make quick decisions, such as whether to avoid the object or continue driving without alteration. Whereas, AVs rely on pre-trained algorithms and sensor data, which may not cover all possible scenarios \citep{pre-trained}. Therefore, this study aims to enhance the autonomous vehicle’s ability to infer and respond to various obstacles it has never encountered before by implementing a KG-based framework within the CARLA autonomous driving simulator through an application programming interface (API) integration \citep{CARLA}. By utilizing graph databases, which excel at representing and processing complex relationships between objects and their physical properties \citep{graph_databases}, our work successfully integrates a KG approach on the sensor-based perception within the CARLA simulator. 

\section{Related Work}

\subsection{Overall Sensors Summary}
First, we examine the function of the three major sensors (cameras, LiDAR, and radars) used in AVs. (i) Cameras produce images of the surrounding environment by detecting light reflected or emitted from objects onto a photosensitive surface. \citep{photosensitive}. Additionally, both moving and stationary obstacles are identifiable by cameras with high-resolution photographs. (ii) LiDAR (Light Detection And Ranging) is a distance-sensing technique that emits laser light pulses or infrared beams toward target impediments. By calculating the time between emission and reception of the light pulse, LiDAR obtains an accurate distance estimate \citep{LIDAR}. LiDAR produces data in the form of point cloud data (PCD), which consists of the (x,y,z) coordinates and the intensity information of the obstacles \citep{PCD}. (iii) Radars determine the relative speed and position of target obstacles using the Doppler shift in electromagnetic waves \citep{Level_5}. As the target approaches the radar, the frequency of the detected signal increases, enabling AVs to determine range information \citep{range}.

\subsection{Problems Associated with Sensors}
Cameras, LiDAR, and radars each have weaknesses when operating independently and also collectively. Cameras show high performance for capturing high-resolution images but struggle to function in poor climatic conditions such as fog or heavy rain and are unable to reliably measure distance \citep{distance}. LiDAR is prone to data distortions such as "black holes" in its point cloud and is not suited for poor weather. Radar performs well in detecting obstacles at long distances and in poor climatic conditions but it is unreliable in distinguishing between complex materials or those with weak echoes, such as plastic or rubber \citep{black_holes}. The issues of unreliability and inaccuracy are amplified when these sensors are combined, a process referred to as sensor fusion, where data from multiple sources are merged for perception \citep{fused}. Current sensor fusion algorithms rely on object-level fusion that involves sensors identifying and classifying objects independently before combining their data. This approach is not optimized as it fails to integrate the strengths of each sensor into a unified collective understanding of the environment \citep{human_drivers}. Furthermore, inconsistencies between sensor inputs can lead to conflicting interpretations which reduces the reliability of AVs. For instance, a lightweight plastic bag might be misclassified as a rigid obstacle due to conflicting data from LiDAR and radar.

\subsection{Convolutional Neural Network Approach}
In response, many studies have focused on embedding deep learning models such as convolutional neural networks (CNNs) and recurrent neural networks (RNNs) for hazardous obstacle detection in AV from dashboard video footage. Particularly, CNNs have shown exceptional performance with (i) image classification (ii) object detection (iii) semantic segmentation, particularly when conditions are well controlled with golden datasets (well-defined datasets) \citep{golden}. Despite great advances from deep learning models like CNNs, unguided learning tends to draw spurious patterns. In addition, CNNs still have limitations when applied to dynamic material recognition and unpredictable road environments. More specifically, while CNNs excel at extracting high-level features from images, they fail to infer materialistic properties such as hardness, elasticity, or malleability that are not differently visible \citep{materials}. Furthermore, CNNs show relatively low performance with inexperienced situations that were not explicitly represented in their training data. Thus, as aforementioned, the lack of relational reasoning when handling obstacles has given rise to a KG-based approach.
\begin{figure}[h]
    \centering
    \includegraphics[width=\linewidth]{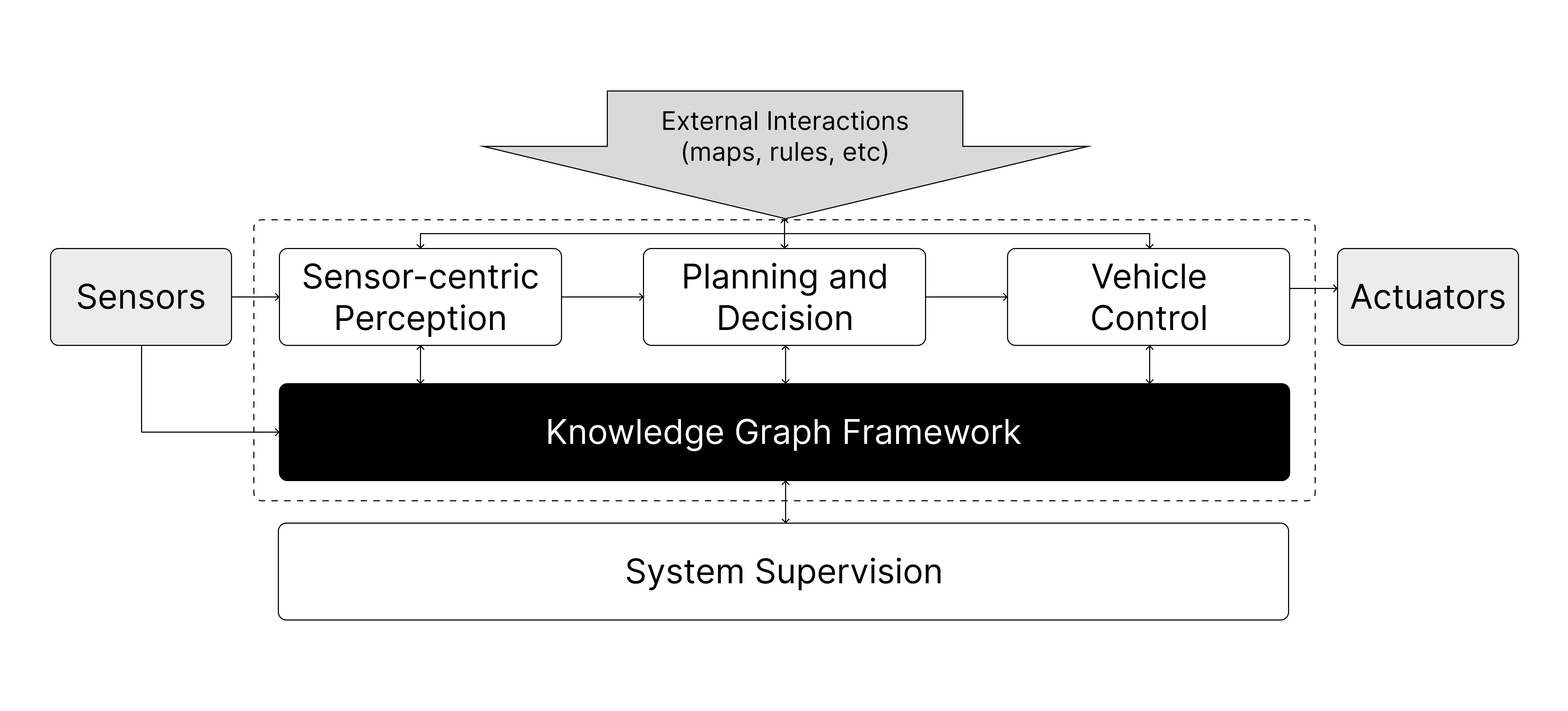}
    \caption{Knowledge Graph Representation of Obstacle Relationships}
    \label{fig:knowledge_graph}
\end{figure}
\subsection{Implementation of Knowledge Graphs}
KGs store and access interrelated data entities as directed labeled multigraphs, enabling semantic reasoning with high accuracy in complex inference tasks \citep{graph_databases}. The study proposes using a KG to represent relationships between obstacles and their materialistic attributes to infer properties beyond what sensors can observe, as shown in Fig.1.

\section{Methodology}
\begin{figure}[h]
    \centering
    \includegraphics[width=\linewidth]{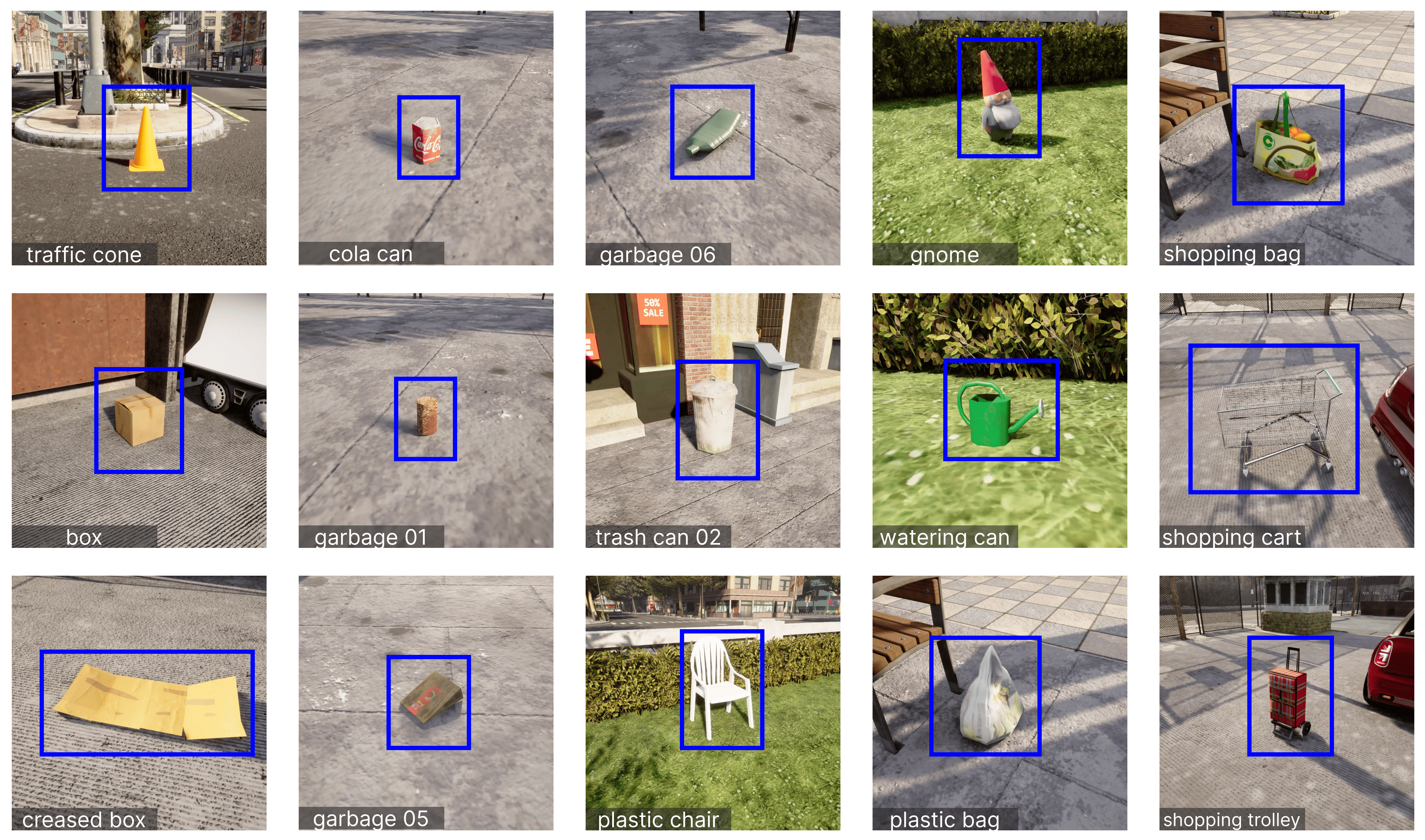}
    \caption{Types of Testing Static Obstacles in CARLA Simulator}
    \label{fig:experimental_setup}
\end{figure}
\subsection{Experimental Design}
In this section, we describe our KG-based integration framework that aims to mimic human inference mechanisms used for material recognition, implemented within CARLA (open-source autonomous driving simulator). This study uses CARLA to control environmental variables and experiment with the impact of obstacle type on vehicle response. To ensure consistency, the following variables are standardized across trials: (i) road conditions (ii) ambient lighting (iii) the behavior of surrounding vehicles. In addition, we examine vehicle behavior within the simulator in two distinct situations: (a) scenarios where lane changes are feasible and (b) scenarios where surrounding vehicles restrict lane changing. 

Furthermore, in our experiments, we conduct a comparative analysis between the AV integrated with the KG framework and the default AV system provided by the CARLA simulator. The default AV exploits programmed algorithms and sensor data to make decisions on handling obstacles within the environment. Whereas, the AV with the KG framework applies semantic reasoning to infer material properties based on input data. For this comparative analysis of distinct AV systems in two scenarios (a) and (b) as mentioned before, fifteen types of static obstacles available from the CARLA simulator were tested, as shown in Fig.2, representing diverse material attributes and real-world relevance. 

The independent variable in this study is the type of obstacle from Fig.2, while the dependent variable is the simulated vehicle’s response (lane change, sudden braking, or driving through). For each scenario, an optimal response exists depending on the material attributes of the object. For instance, for a plastic bag from Fig.2, the optimal action would be to pass through as it is easily deformable. In contrast, for a traffic cone, despite its small dimensions, a lane change would be optimal when surrounding roads are vacant due to its elastic characteristics and the risk of attachment to the tire. Hence, throughout this experiment, the response of the AV integrated with the KG to each obstacle will be examined across several trials.
\subsection{Knowledge Graph Framework Integration}
The primary assumption underlying this experiment is that YOLO (You Only Look Once), used for real-time object detection, will not be utilized \citep{YOLO}. Instead, object detection will be replaced by directly inputting pre-determined object identifiers (e.g., object name and dimensions) into the KG framework. This simplification in the process is justified as the experimental focus is on static obstacles in which the real-time detection capabilities of YOLO are not critical.
\begin{figure}[h]
    \centering
    \includegraphics[width=\linewidth]{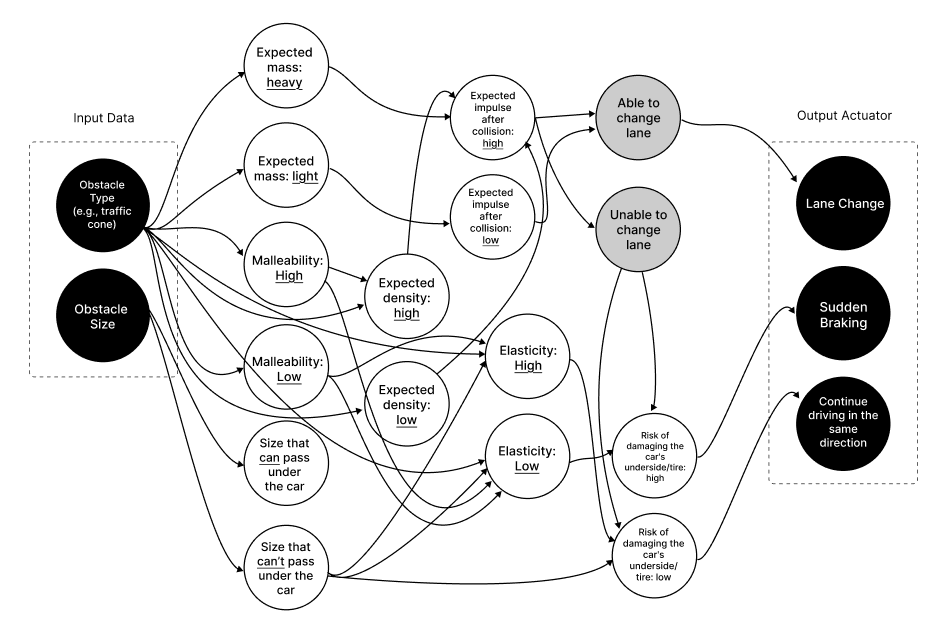}
    \caption{Ontological representation of KG-based integration framework (expressed through nodes, edges, and label)}
    \label{fig:experimental_setup}
\end{figure}
In addition, it prevents the introduction of unnecessary complexity and ensures that the primary objective remains focused. The ontological framework was designed as illustrated in Fig.3, which is composed of three primary layers: input, semantic reasoning, and output. The semantic reasoning is based on two fundamental parameters: obstacle type (e.g., plastic bag, shopping cart, gnome) and obstacle size, which are assumed to be acquired through conventional sensors: cameras, radar, LiDAR, and pre-installed object detection algorithms (e.g., YOLO or an equivalent method). The approach to input two minimal parameters is an attempt to mirror human cognitive processes in driving, as humans also utilize minimal information to make decisions. Subsequently, the framework’s categorical identification associates each obstacle type with a set of predefined material attributes within the KG, including malleability, elasticity, and density, allowing the AV to more fully understand the materialistic characteristics of obstacles. Hence, through the utilization of the semantic reasoning layer, which implements a graph traversal algorithm that processes these inputs through multiple interconnected nodes representing physical properties and risk factors, the vehicle is more likely to reach a comprehensive and consensual conclusion than when using sensors previously. 

Excluding input and output nodes, nodes are created through the following criteria: (i) expected mass: heavy or light, (ii) malleability: heavy or low, (iii) size that can or cannot pass under the car, (iv) expected density: high or low, (v) elasticity: high or low, (vi) expected impulse after collision: high or low, (vii) able or unable to change lane, (viii) risk of exerting strong damage to the car’s underside/tire: high or low. 
\indent Even though there is no universal relationship between each property, (for instance, objects with high malleability do not necessarily always have low density) properties with relatively high correlation are typically connected through edges based on common tendencies. For instance, lesser-density materials typically have lower atomic mass or wider atomic spacings. The free atom movement in malleable materials may be linked to less dense atomic packing, which could result in a lower density. Based on this common logic, relationships have been established between nodes to produce the most optimal conclusion to minimize damage exerted. Moreover, depending on the context, changing lanes may exacerbate the damage, while in some cases sudden braking may increase the risk. Hence, deciding which action to take can be extremely pivotal. By integrating this contextual understanding with material properties and input data, we implemented the KG framework to make the most optimal decision to minimize impact. As a result, complex property inference through relationships is enhanced by this semantic traversal method, while multi-parameter decision optimization is made possible by the KG's logical architecture. The framework was built to maximize reasoning abilities that surpass conventional obstacle avoidance systems by distinguishing between geometrically identical obstacles according to their material attributes. 

On the CARLA simulator, all fifteen objects were tested with and without KG integration for two scenarios: one where lane change was possible and one where it was unsafe. Fig.4 is a sample ontological representation to further explain the mechanism behind the KG-based framework. Using the example of a plastic chair, we conduct a comprehensive comparison of two scenarios: one where lanes can be changed and one where lanes cannot be changed. The plastic chair has the following characteristics: light expected mass; high malleability; a size that cannot pass under the car; low expected density; low elasticity; and low expected impulse after collision (due to low mass and density). Note that in Fig.4a, lane change is available. The logic is as follows: as the mass is light, the expected impulse is low. This is further cross-validated by low density and high malleability, resulting in a consistent conclusion with high reliability. In contrast, in Fig.4b, lane change is not safe, as the condition for plastic chairs remains constant. Thus, the overall expected impulse after the collision is still low but, as shown in Fig.3, the plastic chair may have a high risk of exerting strong damage to the car's underside/tire (e.g., sharp edges of the plastic chair may puncture the tire). Since lane change is not available in Fig.4b, the most optimal actuator output will be sudden braking, assuming that there are no vehicles behind or around. 

\begin{figure}[h]
    \centering
    \begin{subfigure}{0.48\linewidth}
        \centering
        \includegraphics[width=\linewidth]{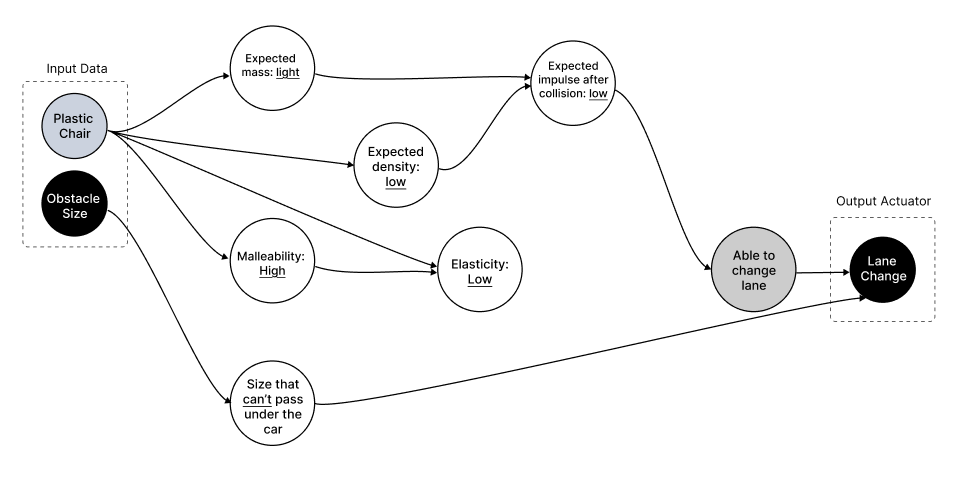}
        \caption{Scenario where lane change is feasible}
        \label{fig:experimental_setup}
    \end{subfigure}
    \hfill
    \begin{subfigure}{0.48\linewidth}
        \centering
        \includegraphics[width=\linewidth]{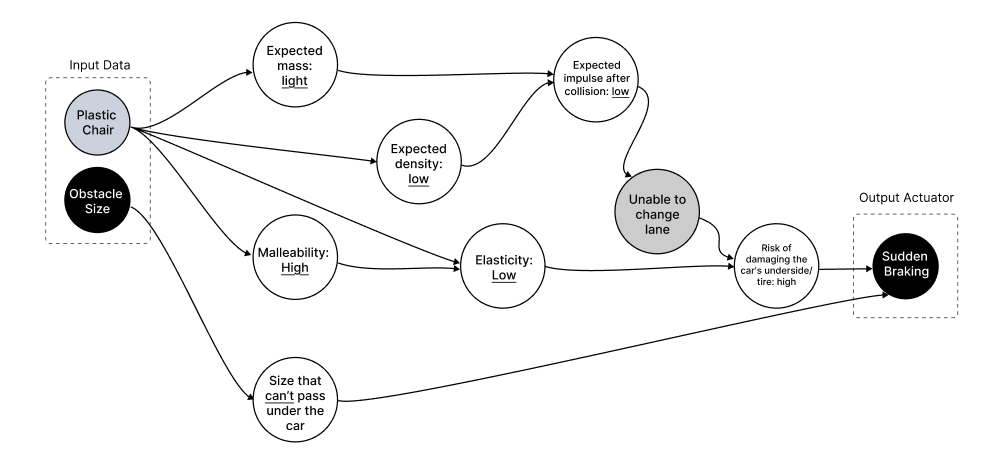}
        \caption{Scenario where lane change is restricted}
        \label{fig:plastic_chair}
    \end{subfigure}
    \caption{KG-based decision-making with a plastic chair}
    \label{fig:side_by_side_figures}
\end{figure}

\subsection{Integration with CARLA simulator}
Through a semantic interoperability framework, an ontology-based KG has been integrated with the CARLA autonomous driving simulator. The KG underwent segmentation to be compatible with Python’s processing constraints, which involves the decomposition of the graph structure into manageable components while maintaining the semantic relationships and ontological hierarchy. Then, the resultant segments are transformed through an intermediary layer that implements data marshaling and semantic mapping. This implementation instantiated a client-server architecture to reconstruct and synchronize the segmented graph components within the simulator. Subsequently, using CARLA’s traffic manager (TM), which constructs realistic urban traffic conditions, we installed real-world relevant obstacles mentioned in Fig.2. Additionally, an autopilot module has been installed in the TM to conduct a comparative analysis of the KG-integrated AV and default autopilot AV. Moreover, autopilot mode has been overridden when obstacles are detected since the perception system of the vehicle becomes unclear when sensors like cameras, LiDAR, and radar present contradicting information. Thus, the system implements a more deterministic and dependable control mechanism by forcing autopilot to stop and move to decision-making based on graph algorithms.
\section{Results}

\begin{table}[h]
\centering
\caption{Qualitative Results of Control and Experimental Groups}
\label{tab:qualitative_results}
\begin{tabular}{lcccc}
\toprule
\textbf{Object} & \multicolumn{2}{c}{\textbf{Lane Change Restricted}} & \multicolumn{2}{c}{\textbf{Lane Change Unrestricted}} \\
\cmidrule(lr){2-3} \cmidrule(lr){4-5}
                & \textbf{Control} & \textbf{Experimental} & \textbf{Control} & \textbf{Experimental} \\
\midrule
Construction cone       & Collide/Stop & Collide/Stop & Lane Change & Lane Change \\
Box 01                  & Collide/Stop & Collide/Stop & Lane Change & Lane Change \\
Creased box 02          & Collide/Stop & Collide/Stop & Collide/Stop & Collide/Stop \\
Cola can                & Collide/Stop & Collide/Stop & Collide/Stop & Collide/Stop \\
Garbage 01              & Collide/Stop & Collide/Stop & Collide/Stop & Collide/Stop \\
Garbage 05              & Collide/Stop & Collide/Stop & Collide/Stop & Collide/Stop \\
Garbage 06              & Collide/Stop & Collide/Stop & Collide/Stop & Collide/Stop \\
Trash can 03            & Collide/Stop & Collide/Stop & Collide/Stop & Lane Change \\
Plastic chair           & Collide/Stop & Sudden Braking & Lane Change & Lane Change \\
Gnome                   & Collide/Stop & Collide/Stop & Collide/Stop & Collide/Stop \\
Watering can            & Collide/Stop & Collide/Stop & Collide/Stop & Lane Change \\
Plastic bag             & Collide/Stop & Collide/Stop & Lane Change & Collide/Stop \\
Shopping bag            & Collide/Stop & Collide/Stop & Lane Change & Lane Change \\
Shopping cart           & Collide/Stop & Sudden Braking & Lane Change & Lane Change \\
Shopping trolley        & Collide/Stop & Collide/Stop & Lane Change & Lane Change \\
\bottomrule
\end{tabular}
\end{table}

\begin{table}[h]
\centering
\caption{Quantitative Results of Control and Experimental Groups}
\label{tab:quantitative_results}
\begin{tabular}{lcccc}
\toprule
\textbf{Metric} & \multicolumn{2}{c}{\textbf{Lane Change Restricted}} & \multicolumn{2}{c}{\textbf{Lane Change Unrestricted}} \\
\cmidrule(lr){2-3} \cmidrule(lr){4-5}
                & \textbf{Control} & \textbf{Experimental} & \textbf{Control} & \textbf{Experimental} \\
\midrule
Total Lane Changes               & 0 (0\%)    & 0 (0\%)     & 7 (46.7\%)  & 8 (53.3\%) \\
Total Sudden Braking Incidents   & 0 (0\%)    & 2 (13.3\%)  & 0 (0\%)     & 0 (0\%)    \\
Total Collide or Stop            & 15 (100\%) & 13 (86.7\%) & 8 (53.3\%)  & 7 (46.7\%) \\
\bottomrule
\end{tabular}
\end{table}

Even though the accident rate of autonomous vehicles is lower than that of human drivers, accidents caused by human mistakes and those caused by machines carry significantly different implications. This study focuses on one of the various risks that may arise during autonomous driving, specifically the situation where obstacles are present in the driving lane and the vehicle must avoid them. Obstacles on the road represent a high-risk situation, as small mistakes can lead to serious accidents. Our purpose is to explore whether combining KGs can provide better avoidance performance than current information processing methods that rely on visual images, geometric data, and motion information from cameras, LiDAR, and radar. This concept is inspired by the method that human drivers avoid obstacles; they do not solely process visual information but also combine it with learned existing knowledge to make decisions.
\subsection{Control group (Default autopilot mode)}

\subsubsection{Lane change restricted}
Through our experimentation, it was validated that integrating KGs could bring improvements in obstacle management. The first control experiment was conducted in a scenario where, upon detecting an obstacle, there were vehicles on adjacent sides. In this scenario, the vehicle did not change lanes or brake suddenly and continued to drive straight ahead, even when encountering obstacles (for all fifteen different obstacle scenarios). It is assumed that the CARLA autopilot driving system was administered to consider a collision with another vehicle in the adjacent lane as a more serious situation than a collision with an obstacle in front. 
\subsubsection{Lane change unrestricted} The second control experiment was conducted in a scenario where lane change was feasible due to the absence of other vehicles in adjacent lanes. In this experiment, lane changes occurred for seven out of the fifteen obstacles. These obstacles were: (1) construction cone, box 01, plastic chair, plastic bag, shopping bag, shopping cart, and shopping trolley. In contrast, lane changes did not occur for the following obstacles: creased box, cola can, garbage 01, garbage 05, garbage 06, gnome, and watering can. Although there was no clear distinction between the group of obstacles that led to lane changes and the group that did not, one potential explanation could be that the obstacles in the CARLA simulator were not constructed with the same size, shape, and properties as the ones seen in real-life footage. However, it's important to note that the obstacles for which lane changes occurred, such as the box, plastic chair, shopping cart, and shopping trolley were larger than the obstacles that did not cause lane changes, such as the creased box, garbage, and cola can. 

\begin{figure}[h]
    \centering
    \includegraphics[width=\linewidth]{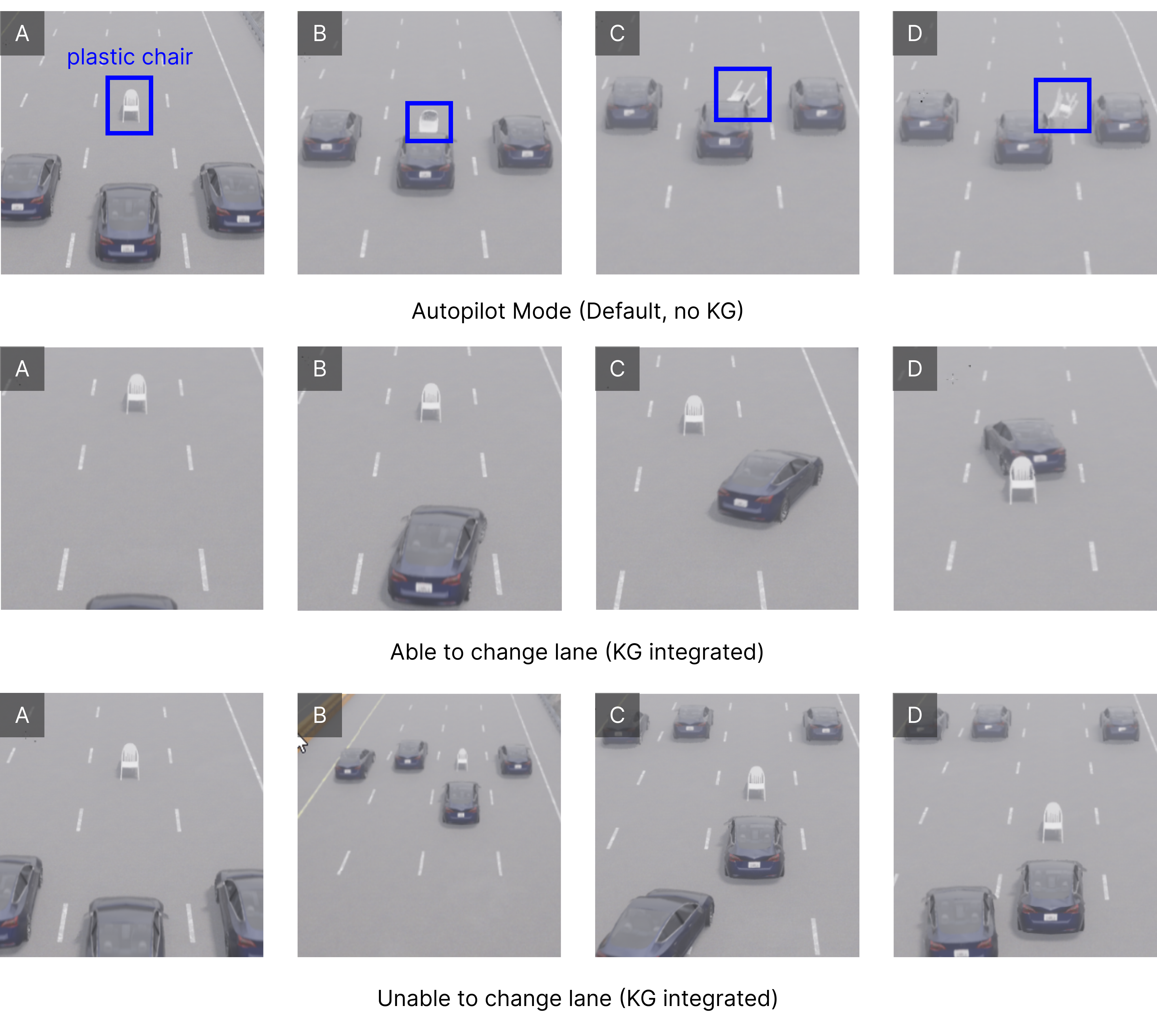}
    \caption{Scenarios with autopilot mode (default, no KG) / able to change lane (KG integrated) / unable to change lane (KG integrated)}
    \label{fig:experimental_setup}
\end{figure}

\subsection{Experimental Group (KG-integrated mode)}
\subsubsection{Lane change restricted}
The investigation for the experimental group was conducted in two parts, similar to the control group. The first scenario for Experimental Group 1 involved fifteen obstacles. In this experiment, the vehicle encounters an obstacle where a car is present in the adjacent lane, making lane changes difficult. As mentioned previously, when the vehicle encounters an obstacle, autopilot mode is deactivated and the obstacle is handled based on the decision-making process of the graph algorithm. In this experiment, for two obstacles, a plastic chair and a shopping cart, the vehicle performs an emergency stop, while for the other thirteen obstacles, the vehicle continues driving straight without stopping. 

On the graph ontology, the plastic chair and shopping cart share similar characteristics: they are sized in a way that makes it difficult for the vehicle to pass under them, and they are expected to produce significant impact force upon collision. Additionally, they have high elasticity, meaning that when a collision occurs, they maintain their shape and may get stuck under the vehicle. More specifically, obstacles like plastic chairs and shopping carts, which have high elasticity and strength, can bounce upon impact and cause accidents with nearby vehicles, making them incredibly dangerous. 

In contrast, the gnome is categorized as a doll, typically expected to have a low impact force and not significantly affect the vehicle’s underside or wheels upon collision. However, the results from the CARLA simulation reveal that when the vehicle collided with the gnome, its direction was altered, almost leaving its lane. The reason for this unexpected outcome is that, unlike what was reflected in the ontology, the gnome was made of a very hard and dense material. As a result, when the vehicle collided with the gnome, it remained intact and became lodged underneath the vehicle, interfering with the vehicle’s undercarriage, as illustrated in Fig. 7.

\begin{figure}[h]
    \centering
    \includegraphics[width=\linewidth]{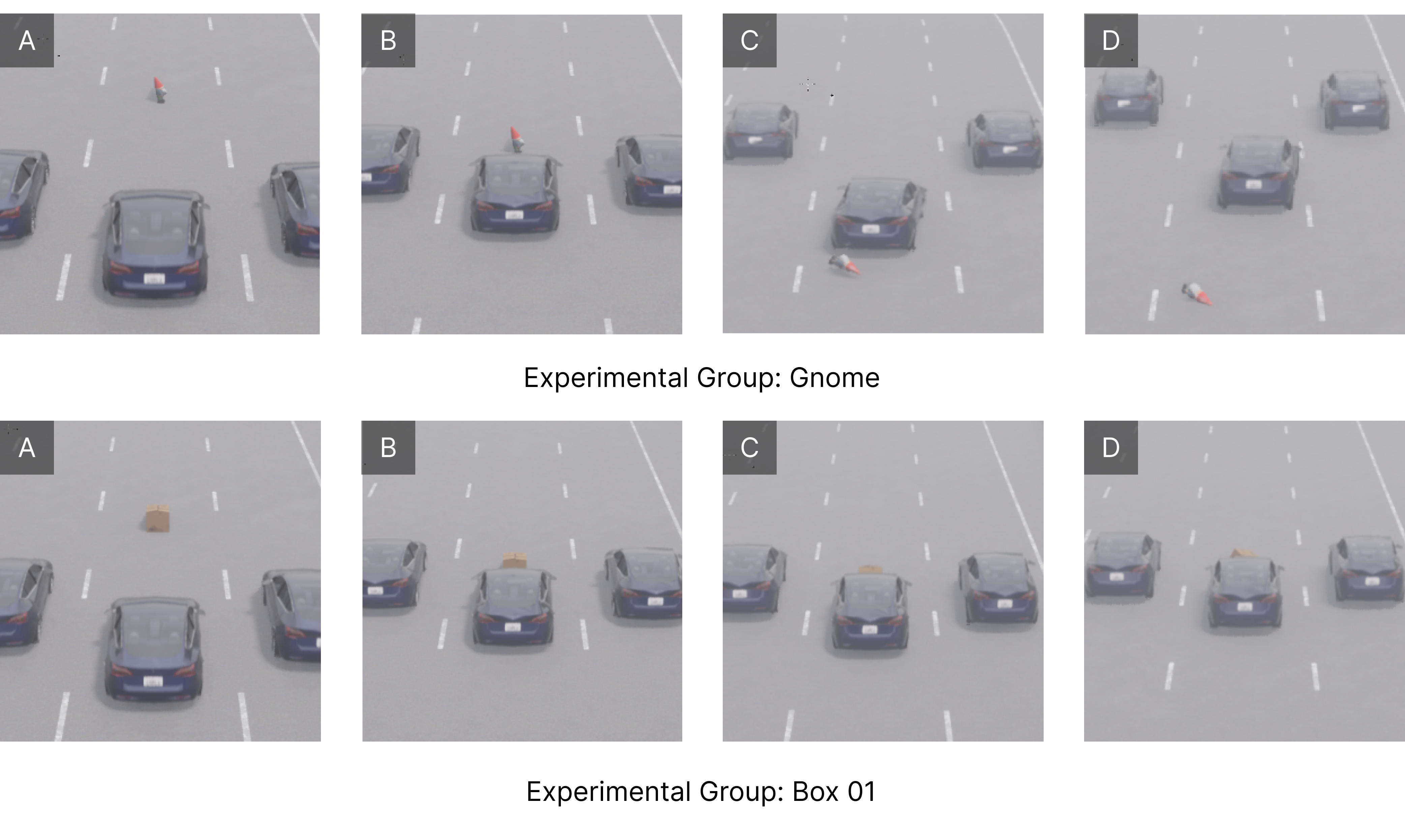}
    \caption{Scenarios with gnome and box 01 (experimental group)}
    \label{fig:experimental_setup}
\end{figure}

\subsubsection{Lane change unrestricted}
The second experimental group involved cases where lane changes were enabled. In this group, lane changes were made for eight out of the fifteen obstacles: construction cone, box 01, trash can 03, plastic chair, watering can, shopping bag, shopping cart, and shopping trolley. Compared to the control group in Experiment 2, where lane changes were feasible, twelve of the same obstacles overlapped, and lane changes were made for two additional obstacles: the trash can and the watering can.

\subsection{Final Observations}
In the control group where lane changes were restricted, the default autopilot mode collided with obstacles or stopped for 100\% of the trials (n = 15). In scenarios where lane changes were enabled, the autopilot changed lanes in 46.7\% of cases and for the remaining 53.3\%, it either collided with obstacles or stopped. In the experimental group where a KG framework was implemented, the system performed a sudden brake for 13.3\% of cases when lane changes were enabled and collided or stopped for 86.7\% of the remaining cases. Under conditions where lane change was restricted, 53.3\% of decisions resulted in a lane change, whereas 46.7\% led to either a collision or a stop. Overall, the experimental results with the KG framework demonstrated significant quantitative improvements in obstacle response strategies, particularly in scenarios that required specialized and nuanced reasoning of material properties. While the default autopilot system collided 100\% of the time in scenarios where lane changes were restricted, the KG integrated framework provoked emergency stops for plastic chairs and shopping carts. (13.3\% cases) Furthermore, the KG approach exhibited optimized lane change decisions, increasing lane changes by 6.6\% (8/15 vs 7/15 in controls) for garbage 05 and trash cans. Importantly, it also avoided unnecessary lane changes for plastic bags, correctly deciding to drive through deformable obstacles. Ultimately, the KG outperformed sensors by prioritizing material properties, such as elasticity and density, over solely considering geometric data.
\section{Conclusion}
Despite the remarkable advancements in autonomous driving technology, occasional reports of major accidents caused by misjudgments of AVs continue to raise concerns. Although the accident rate of autonomous vehicles is lower than that of human drivers, the nature of accidents caused by both scenarios is very distinct \citep{waymo}. Autonomous driving technology requires extremely high technical perfection; even though the chance of an accident occurring is tiny, the severity of the event could be fatal. Currently, AV technology has not been demonstrated to meet or surpass human driving abilities in all scenarios. This study addresses the hypothesis that combining sensors with a KG framework would improve AVs' comprehension of physical material qualities. Simply increasing data collection and processing may not be sufficient enough to exceed the processing capabilities of human drivers, who rely on both knowledge and experience. Instead, achieving higher-level autonomous driving may demand the integration of other types of criteria such as knowledge-based reasoning and experience alongside data-centric approaches. Using the CARLA simulator, we compared experiments with and without KG integration in the obstacle avoidance process. Our findings demonstrate that the KG-enabled system executed more sophisticated obstacle avoidance than the autopilot mode in situations such as emergency braking ahead of a plastic chair or shopping cart and changing lanes in front of a garbage can or watering can. Quantitatively, the KG framework performed improved decision-making by 13.3\% of cases when lane changes were restricted and 6.6\% when lane changes were open, indicating the potential for KG's to improve obstacle response strategies beyond sensor approaches.
\section{Acknowledgments}
We would like to thank Professor Leilani H. Gilpin at the University of California, Santa Cruz, for her valuable feedback and guidance. Additionally, we acknowledge the use of GPUs and compute resources provided by the AIEA Lab for running our experiments on the CARLA simulator.


\bibliography{iclr2025_conference}
\bibliographystyle{iclr2025_conference}

\end{document}